\pdfoutput=1

\documentclass[11pt]{article}

\usepackage[preprint]{acl}

\usepackage{times}
\usepackage{latexsym}
\usepackage{mathtools}
\usepackage{booktabs}
\usepackage{adjustbox}

\usepackage{multirow}

\usepackage[T1]{fontenc}

\usepackage[utf8]{inputenc}

\usepackage{microtype}

\usepackage{inconsolata}

\usepackage{graphicx}

\usepackage{caption}
\usepackage{subcaption}
\usepackage{float}
\restylefloat{table}
\usepackage{makecell}
\usepackage{ucs}
\usepackage{csvsimple}

\usepackage{amssymb,graphicx,stackengine,xcolor}
\def\ucr{\scalebox{1}{\stackinset{c}{}{c}{-.2pt}{%
  \textcolor{white}{\sffamily\bfseries\small ?}}{%
  \rotatebox{45}{$\blacksquare$}}}}

\usepackage{amsmath}
\newcommand*{\crosssymbol}{%
    \text{%
      \raisebox{1ex}{%
        \makebox[0pt][l]{%
          \rule[-.2pt]{.75ex}{.4pt}%
        }%
        \makebox[.75ex]{%
          \rule[-1ex]{.4pt}{1.5ex}%
        }%
      }%
    }%
}

\usepackage{ifthen}
\newcommand*{\GottBERT}[2]{\ifthenelse{\equal{#2}{last}}{$\mathrm{GottBERT}_{#1}^{\crosssymbol}$}{$\mathrm{GottBERT}_{#1}$}}
\newcommand*{\GottBERTf}[2]{\ifthenelse{\equal{#2}{last}}{$^{\mathrm{f}}\mathrm{GottBERT}_{#1}^{\mathrm{\crosssymbol}}$}{$^{\mathrm{f}}\mathrm{GottBERT}_{#1}$}}

\newcommand{\GottBERTt}{$\mathrm{GottBERT}$}

\title{GottBERT: a pure German Language Model}

\author{
        \textbf{Raphael Scheible\textsuperscript{\,{\normalfont 1}}},
        \textbf{Johann Frei\textsuperscript{\,{\normalfont 2}}},
        \textbf{Fabian Thomczyk\textsuperscript{\,{\normalfont 3}}},
        \textbf{Henry He\textsuperscript{\,{\normalfont 4}}},
        \textbf{Patric Tippmann\textsuperscript{\,{\normalfont 5,6}}},\\
        \textbf{Jochen Knaus\textsuperscript{\,{\normalfont 5}}},
        \textbf{Victor Jaravine\textsuperscript{\,{\normalfont 7}}},
        \textbf{Frank Kramer\textsuperscript{\,{\normalfont 2}}} and 
        \textbf{Martin Boeker\textsuperscript{\,{\normalfont 1}}} \\\\
  \small\textsuperscript{\,{\normalfont 1}} Institute for AI and Informatics in Medicine, University Hospital rechts der Isar, Technical University Munich\\
  \small\textsuperscript{\,{\normalfont 2}} IT-Infrastructure for Translational Medical Research, Faculty of Applied Computer Science, University of Augsburg\\
  \small\textsuperscript{\,{\normalfont 3}} Data Integration Center, Faculty of Medicine, University of Freiburg\\
  \small\textsuperscript{\,{\normalfont 4}} School of Computation, Information and Technology, Technical University Munich\\
  \small\textsuperscript{\,{\normalfont 5}} Institute of Medical Biometry and Statistics, Medical Center, Faculty of Medicine, University of Freiburg\\
  \small\textsuperscript{\,{\normalfont 6}} Freiburg Center for Data Analysis and Modeling, University of Freiburg\\
  \small\textsuperscript{\,{\normalfont 7}} Hengrui Europe Biosciences, Zurich
\\
\\
 \small{
   \textbf{Correspondence:} \href{mailto:raphael.scheible@tum.de}{raphael.scheible@tum.de}
 }
}

\begin{document}
\maketitle
\begin{abstract}
Pre-trained language models have significantly advanced natural language processing (NLP), especially with the introduction of BERT and its optimized version, RoBERTa. While initial research focused on English, single-language models can be advantageous compared to multilingual ones in terms of pre-training effort, overall resource efficiency or downstream task performance.
Despite the growing popularity of prompt-based LLMs, more compute-efficient BERT-like models remain highly relevant.
In this work, we present the first German single-language RoBERTa model, {\GottBERTt}, pre-trained exclusively on the German portion of the OSCAR dataset. Additionally, we investigated the impact of filtering the OSCAR corpus.
{\GottBERTt} was pre-trained using fairseq and standard hyperparameters. We evaluated its performance on two Named Entity Recognition (NER) tasks (Conll 2003 and GermEval 2014) and three text classification tasks (GermEval 2018 fine and coarse, and 10kGNAD) against existing German BERT models and two multilingual models. Performance was measured using the $F_{1}$ score and accuracy.
The {\GottBERTt} base and large models showed competitive performance, with {\GottBERTt} leading among the base models in 4 of 6 tasks. Contrary to our expectation, the applied filtering did not significantly affect the results.
To support the German NLP research community, we are releasing the {\GottBERTt} models under the MIT license.
\end{abstract}

\section{Introduction}
The computation of contextual pre-trained word representations is the foundation of neural language modeling (LM) in natural language processing (NLP).
The field of NLP experienced remarkable progress by the use of transformer-based approaches \cite{vaswani_attention_2017}. Especially Bidirectional Encoder Representations from Transformers (BERT) \cite{devlin_bert_2019} impacted the field which subsequently was robustly optimized to RoBERTa \cite{liu_roberta_2019}. 
These transformer-based approaches rely on large-scale pre-trained language models, which are subsequently fine-tuned through supervised training on specific downstream tasks, leveraging the context representations learned from the generic domain to achieve superior performance compared to training from scratch, a process known as transfer learning.
On the other hand, the computation of the language model is performed self-supervised. Large text blobs are required for training and strong hardware such as hundreds of Graphics Processing Units (GPU) \cite{martin_camembert_2020} or Tensor Processing Units (TPU) \cite{you_large_2020}. Initially, most of the research took place in English followed by multilingual approaches \cite{conneau_unsupervised_2019, NEURIPS2019_c04c19c2}. Although, multilingual approaches were trained on large texts of many languages, they can be outperformed by single language models \cite{de_vries_bertje_2019,martin_camembert_2020, le_flaubert_2020, delobelle_robbert_2020}. Additionally, a single language model requires fewer computational resources and a smaller dataset compared to the vast and varied data needed for multilingual models. Single language models trained with the Open Super-large Crawled ALMAnaCH coRpus (OSCAR) \cite{ortiz_suarez_monolingual_2020} showed good performance due to the size and variance of the OSCAR corpus \cite{martin_camembert_2020,delobelle_robbert_2020}.
The focus of pre-training language models has shifted towards scaling up transformer-based large language models (LLMs) \cite{touvron2023llama1, touvron2023llama2, Liu_2023, jiang2023mistral}. These models, like Llama 3\footnote{https://github.com/meta-llama/llama3}, are vastly larger than the aforementioned models \cite{touvron2023llama1, touvron2023llama2}. Despite the advantages of LLMs, such as gradient-free prompting, smaller models remain valuable for their efficiency and practical deployment.
For these reasons, we pre-trained the first German RoBERTa single language models with the German portion of the first published deduplicated version of OSCAR - the German OSCAR text trained BERT ({\GottBERTt}). Inspired by FlauBERT \cite{le_flaubert_2020}, we also trained models with a filtered version of OSCAR.
In an evaluation we compared the performance of all models on the two named entity recognition tasks Conll 2003 and GermEval 2014, NLI as well as on the text classification tasks GermEval 2018 and GNAD with existing German single language BERT and  models and two multilingual models.

Our contributions can be summarized as follows:
\begin{itemize}
    \item We introduced a filtering method specifically applicable to German texts which we applied to the first version of the German portion of the OSCAR corpus.
    \item We pre-trained single language RoBERTa models specifically for the German language based on the filtered and original OSCAR corpus. These models are publicly available under the MIT open-source license.
    \item We evaluated the models on five downstream tasks (3 classification, 2 NER and NLI). Further, we demonstrated the effects of training the model with the filtered corpus.
\end{itemize}
 
\section{Related Work}

Most recently, transformer-based models widely impacted the field of NLP. From neural translation \cite{ott_scaling_2018, ng_facebook_2019} to generative language models starting with GPT-2 \cite{radford_language_2019}, remarkable performance gains were achieved. With BERT, an approach to facilitate pre-trained transformer-based models was introduced. Fine-tuned on downstream tasks, BERT-based approaches improved the performance of several NLP tasks \cite{devlin_bert_2019,liu_roberta_2019}. However, BERT models were first released as single-language models in English based on 16GB of raw text and as the multilingual model mBERT based on Wikipedia in about 100 languages \cite{devlin_multilingual_2018}. These models were followed by single-language models for several languages: Bertje \cite{de_vries_bertje_2019} for Dutch, FinBERT \cite{virtanen_multilingual_2019} for Finish, GermanBERT \footnote{\url{https://deepset.ai/german-bert}} and a German BERT from the MDZ Digital Library team at the Bavarian State Library to which we refer to as dbmz BERT in this paper\footnote{\url{https://github.com/dbmdz/berts\#german-bert}}.
GermanBERT was trained using 12GB of raw text data basing on the German Wikipedia (6GB), the OpenLegalData dump (2.4GB) and news articles (3.6GB). dbmz BERT used as source data a German Wikipedia dump, EU Bookshop corpus, Open Subtitles, CommonCrawl, ParaCrawl and News Crawl which sums up to a dataset of 16GB.
With the release of RoBERTa a new standard for raw text size was set as it was trained on 160GB of raw English text. Further, RoBERTa enhances the original BERT approach by removing segment embeddings, next sentence prediction and improved hyperparameters. Additionally, instead of using wordpiece \cite{schuster_japanese_2012} tokenization, RoBERTa utilizes GPT2's byte pair encoding (BPE) \cite{radford_language_2019} with the benefit that language-specific tokenizers are not required. Other than mBERT, the multilingual XLM-RoBERTa \cite{conneau_unsupervised_2019} was trained on 2.5TB of filtered CommonCrawl data. CamemBERT is a French RoBERTa model that was trained on the OSCAR and uses sentencepiece \cite{kudo_sentencepiece_2018} BPE. Further, they pre-trained a model with 4GB of the French OSCAR portion and another model with 4GB of the French Wikipedia. The comparison of these models using downstream tasks shows that high text variance leads to better results. UmBERTo\footnote{\url{https://github.com/musixmatchresearch/umberto}} is an Italian RoBERTa model, similarly designed as CamemBERT.
RobBERT, the Dutch single language RoBERTa, was trained on 39GB of the Dutch portion of the OSCAR and outperformed Bertje. A more recent version of RobBert showed the performance gains of language specific BPE compared to the English based GPT2 BPE in downstream tasks.
Susequently, FlauBERT \cite{le_flaubert_2020} for French was released trained on 71GB data. They cleaned a 270GB corpus of mixed sources by filtering out meaningless content and Unicode-normalization. Data was pre-tokenized by moses \cite{koehn_moses_2007} and encoded by fastBPE\footnote{\url{https://github.com/glample/fastBPE}} which is an implementation of \citet{sennrich_neural_2016}.
After the publication of RoBERTa, Google released ELECTRA \cite{clark2020electra} which denoted an improvement to the BERT architecture. Based on these developments, further German language models were then published: GBERT and GELECTRA \cite{chan-etal-2020-germans}. These models were trained on the German portion on OSCAR (145GB) besides three other datasets (18.4GB) and outperformed previously released German single language models.

\section{Methodology}
Following the approach of utilizing the OSCAR, we computed the German OSCAR text trained BERT (\GottBERTt). However, the drawback of BERT approaches is the computational power requirement. Multiple GPUs or TPUs were used for pre-training. All previously listed RoBERTa-based models were computed on GPUs whereas {\GottBERTt} is the first published RoBERTa model pre-trained on TPUs.

\subsection*{Training Data}
The {\GottBERTt} model is trained on the German portion of the OSCAR, a large multilingual text corpus extracted from Common Crawl. The German data portion of the first published deduplicated version of the OSCAR measures 145GB of text containing approximately 21.5 billion words in approximately 459 million documents (one document per line).

\subsubsection*{Filtering OSCAR}
While screening the German OSCAR portion, some issues attracted our attention:
\begin{enumerate}
    \item erroneous umlauts
    \item meaningless documents such as spam, e.g. lists of words
    \item non-German documents
\end{enumerate}

We were able to trace back the cause of wrong umlauts to decoding errors. According to our findings, when an umlaut is considered to be a non-UTF8 encoding, but actually is already UTF8, wrong characters are generated (see Table \ref{tab:umlauts}). In other cases, where the encoding wasn't reproducible, the sign \ucr, is shown. Consequently, lines with at least one \ucr\ were removed. To the rest of documents, we corrected the encoding by applying clean-text\footnote{https://github.com/jfilter/clean-text}. The tool was further configured to remove phone numbers, email addresses, URLs and emojis. Also only documents with a length of at least 40 characters were considered. Secondly, we applied a language detection algorithm which especially filtered lines belonging to ASCII arts (see Apendix \ref{sec:OSCAR}). Due to the corpus size, an efficient Rust implementation\footnote{https://github.com/greyblake/whatlang-rs} of a language detection based on n-gram based text categorization was used \cite{cavnar_n-gram-based_1994}. 

Finally, also due to the corpus size, we trained and applied a single-class SVM \cite{scholkopf1999single} which was trained to filter meaningless documents. In this respect, a special feature of the German language is the capitalization of nouns. In the 17th and 18th century there was a trend in the English and Swedish languages to write nouns with an initial capital \cite{crystal_cambridge_2003,solling_sma_2009}. Dutch had this rule until a spelling reform in 1948. German kept this special rule, although undergoing several orthography reforms in the 20th century. However, Germans sometimes tend to neglect this rule, especially in social media. Therefore, social media platforms might not be a good source for texts for research requiring good quality of orthography, although these texts lead to admirable results in the generative model GPT-2 \cite{radford_language_2019}. In the German OSCAR portion, it was noticeable that documents with little meaning were often written completely in capital letters, had many punctuation marks in relation to words, had a lot of nouns or no stop words at all. Based on this knowledge, the following ratios were computed for each document $D$ consisting of tokens $t_0, t_1, ... , t_{n-1}, t_n$:
\begin{itemize}
    \item stopword ratio:
    \[r_s = \frac{\sum_{i}^{n}\sum_{t_s \in S}[t_i = t_s]}{n},\]
    
    where
    $S$ is the set of stop word tokens for which we used nltk's stopword list.
    
    \item punctuation ratio:
    \[r_p = \frac{|\{\,t_i\,|\, t_i \notin W(D)\,\}|}{n},\]
    where $W(D)$ are the word tokens $w_0, w_1, .. , w_{m-1}, w_m$ of $D$.

    \item unique words ratio:
    \[r_u = \frac{|\{\,w_i\,|\, \forall w_j \in W(D): w_i \neq w_j\}|}{n},\]

    \item upper token ratio:
    \[r_{up} = \frac{\sum_{w_i \in W(D)}c(w_i)}{n},\]
    where 
    \[
        c(w) =
        \begin{cases*}
          1     & if w is capital \\
          0     & otherwise
        \end{cases*}
    \]

\end{itemize}

\inputencoding{utf8x}
\begin{table*}[ht]
\centering
\begin{tabular}{lccccccc}
\hline Special Character & \"a  & \"u  & \"o  & \ss  & Ä  & Ü  & Ö  \\ \hline 
ISO-8859-1   & Ã¤ & Ã¼ & Ã¶ & Ã[U+009F] & Ã[U+0084] & Ã[U+009C] & Ã[U+0096] \\
ISO-8859-2   & Ă¤ & Ăź & Ăś & Ă[U+009F] & Ă[U+0084] & Ă[U+009C] & Ă[U+0096] \\
ISO-8859-4   & Ã¤ & Ã[U+0167] & Ãļ & Ã[U+009F] & Ã[U+0084] & Ã[U+009C] & Ã[U+0096] \\
ISO-8859-9   & Ã¤ & Ã¼ & Ã¶ & Ã[U+009F] & Ã[U+0084] & Ã[U+009C] & Ã[U+0096] \\
ISO-8859-10  & ÃĪ & Ãž & Ãķ & Ã[U+009F] & Ã[U+0084] & Ã[U+009C] & Ã[U+0096] \\
ISO-8859-16  & Ă€ & ĂŒ & Ă¶ & Ă[U+009F] & Ă[U+0084] & Ă[U+009C] & Ă[U+0096] \\
Windows-1250 & Ă¤ & ĂĽ & Ă¶ & Ăź & Ă„ & Ăś & Ă– \\
Windows-1252 & Ã¤ & Ã¼ & Ã¶ & ÃŸ & Ã„ & Ãœ & Ã– \\
\hline 
\end{tabular}
\caption{This table shows the result of wrong encoding of German umlauts. Artifacts occur whenever a file is expected to be encoded by an appropriate encoding, but truly is UTF-8 encoded.}\label{tab:umlauts}
\end{table*}
\inputencoding{utf8}

To train the single class SVM, 12k documents were used for the unsupervised training. The performance was measured based on 1750 documents annotated documents consisting of two classes: 334 dirty (15.5\%) and 1466 clean (81.5\%). Documents classified as dirty were content of weak meaning, including lists of words, source code, documents mainly using ASCII signs and documents lacking of spaces or having wrong spacing between words. The single hyperparameter $nu$ was optimized by a grid search. The best performing SVM had a weighted $F_1$-score of 0.8578 and 0.5663 as Matthews correlation coefficient \cite{chicco_advantages_2020,chicco_matthews_2021}.
After filtering, the corpus measured 121GB of text containing approximately 18.1 billion words in approximately 382 million documents (one document per line).

\subsection*{Pre-processing}
Originally, RoBERTa uses GPT-2 \cite{radford_language_2019} byte pair encoding to segment the input into subword units. Therefore, no pre-tokenization is required and thus no language-specific tokenizer as e.g. moses \cite{koehn_moses_2007} must be used. Its original vocabulary was computed on English data. For {\GottBERTt} we computed a vocabulary of 52k subword tokens based on 40 GB randomly sampled documents of the German OSCAR portion. Compared to the original GPT-2 tokenizer, which was trained on English data, this leads to a 40\% smaller size of the binary data which are fed into fairseq \cite{ott_fairseq_2019}. Furthermore, according to \citet{delobelle_robbert_2020}, it leads to a performance increase. 

\subsection*{Pre-training}
Using fairseq, we pre-trained the {\GottBERT{base}{x}} model using the unfiltered OSCAR on a 256 core TPUv3 pod. The remaining {\GottBERTt} models were computed on a 128 core TPUv4 \cite{jouppi_tpu_2023} pod. We trained the models with RoBERTa base architecture in 100k update steps using a batch size of 8k. A 10k iteration warmup of the learning rate to a peak of 0.0004 was applied, from which the learning rate polynomially decayed to zero. The models with RoBERTa large architecture were trained with the same properties but a peak learning rate of 0.00015. After training on both the filtered and unfiltered OSCAR datasets, we developed four models: {\GottBERT{base}{x}} and {\GottBERT{large}{x}} using the unfiltered as well as {\GottBERTf{base}{x}} and {\GottBERTf{large}{x}} using the filtered dataset. Further, we evaluated each epoch and saved its checkpoint, potentially leading to multiple checkpoints per model setup, namely best and last. The latter ones are indicated with a $\crosssymbol$, e.g. {\GottBERT{base}{last}}. The base models took ca. 1.2 days computation time, while the large ones computed ca. 5.7 days.

\subsection*{Downstream Tasks}
Based on the pre-trained BERT models, several downstream tasks were trained. The training was conducted using the scripts provided by Huggingface \cite{wolf_huggingfaces_2020}. Hyperparameter optimization was performed through a grid search focusing on batch size and learning rate. We trained the downstream tasks NER and CLS with a maximum of 30 epochs.

For natural language inference (NLI), we utilized the hyperparameters specified by Facebook (originally implemented in Fairseq), adopting them to the extent they were available within the Huggingface framework. These tasks were trained with a maximum of 10 epochs.

In order to evaluate the performance, each downstream task ran 24 times using different batch sizes and learning rates. To determine the best checkpoint after training, we select the checkpoint that yields the best $F_{1}$ scores (accuracy for NLI) on the evaluation set. The score is the best of 24 runs of the respective experiment of each trained model. The best score selection is based on the validation set. In terms of performance, our models were compared with six other models listed in Table \ref{bert-table}.

\begin{table*}[h!tbp]
\begin{center}
\begin{tabular}{lcccc}
\hline \bf Model & \bf Type & \bf \#lang & \bf Data Size & \bf Data Source \\ \hline
{\GottBERTt}                    & RoBERTa & 1 & 145GB & \makecell[l]{OSCAR} \\
{\GottBERTf{}{}}                   & RoBERTa & 1 & 121GB & \makecell[l]{filtered OSCAR} \\
GBERT                       & BERT & 1 & 163.4GB & \makecell[l]{OSCAR, OPUS,\\ Wikipedia, OpenLegalData}\\
GELECTRA                    & ELECTRA & 1 & 163.4GB & \makecell[l]{OSCAR, OPUS,\\ Wikipedia, OpenLegalData}\\
dbmz BERT                   & BERT & 1 & 16GB & \makecell[l]{Wikipedia, EU Bookshop corpus\footnote{http://opus.nlpl.eu/EUbookshop.php}, \\ Open Subtitles, \\ Common-,Para-,NewsCrawl} \\ 
mBERT\textsubscript{cased}  & BERT & 104 & \makecell{unknown} & \makecell[l]{Wikipedia} \\
GermanBERT                 & BERT & 1 & 12GB & \makecell[l]{news articles, Open Legal Data\footnote{http://openlegaldata.io/research/2019/02/19/court-decision-dataset.html},\\  Wikipedia} \\
XLM RoBERTa                 & RoBERTa & 100 & \makecell{2.5TB \\ (66.6GB German)} & \makecell[l]{CommonCrawl, Wikipedia} \\
\hline
\end{tabular}
\end{center}
\caption{\label{bert-table} This table shows the models, we used in our experiments. Additional information about the pre-training and architecture is listed. \#lang is the number of languages. Unfortunately, for mBERT we did not find any estimate about the data size.}
\end{table*}

\paragraph{NLI}
NLI entails predicting whether a hypothesis sentence is entailed by, neutral towards or contradicts a premise sentence. We assessed our model on NLI using the German portion of the XNLI dataset \cite{conneau-etal-2018-xnli}. The XNLI dataset is an extension of the Multi-Genre NLI (MultiNLI) corpus \citet{williams-etal-2018-broad}, expanded to 15 languages by manually translating the validation and test sets into each language. For languages other than English, the training set is machine translated. The dataset includes 122k training examples, 2490 development examples, and 5010 test examples for each language. Typically, NLI performance is measured using accuracy.

\paragraph{Named Entity Recognition}
We evaluated {\GottBERTt} on two NER tasks. One was the German part of CoNLL 2003 shared task \cite{tjong_kim_sang_introduction_2003}. It contains three main entity classes and one for other miscellaneous entities. As measurement we used the \emph{harmonic mean of precision and recall} $F_{1}$.
The second NER task was GermEval 2014 \cite{benikova_germeval_2014}. It extends the CoNLL 2003 shared task by fine-grained labels and embedded markables. Fine-grained labels allow the indication of NER subtypes common in German, namely derivations and parts: e.g. ``Mann'' $\rightarrow$ ``männlich'' and ``Mann'' $\rightarrow$ ``mannhaft''.
In order to recognize nested NEs embedded markables are required. Specifically, this was realized by annotating main classes as well as two levels of subclasses. 
Performance was measured by the use of an adapted $F_{1}$ evaluation metric \newcite{benikova_germeval_2014}, which considers the equality of labels and spans (text passages) and additionally levels in the class hierarchy.

\paragraph{Text Classification}
GermEval task 2018 \cite{risch_fine-grained_2018} is a text classification task that contains two subtasks of different granularity: the coarse-grained binary classification of German tweets and fine-grained classification of the same tweets into four different classes. As this datasets does not provide a pre-defined validation set, we used 54\% of the original training set for training, 6\% for validation and 40\% for test.
With this split decision, we sticked to \citet{chan-etal-2020-germans}.
Based on the One Million Posts Corpus \cite{schabus_one_2017}, 
the 10k German News Articles Dataset (10kGNAD) topic classification benchmark\footnote{\url{https://huggingface.co/datasets/community-datasets/gnad10}} was created. The dataset contains approximately 10k news articles from an Austrian newspaper which are to be classified into 9 categories. Usually, 10kGNAD does not provide a pre-defined split. However, the version we used provides a split, using 90\% of the original set for training and 10\% for test. We split 10\% from the training set for validation. For evaluation of both tasks we computed the mean of the $F_{1}$-scores of each class/category.

\section{Results}
As {\GottBERT{large}{x}} was the same checkpoint for last and best, we ended up with 7 {\GottBERTt} checkpoints. {\GottBERT{base}{best}} was saved after 91848 training steps (12 epochs). The filtered models, both {\GottBERTf{large}{best}} and {\GottBERTf{base}{best}}, were saved also saved  after they trained 94530 steps (15 epochs). For these models, this was approximately 1 epoch earlier then the full training steps. The dirty models trained up to 13.07 epochs and the filtered ones up to 15.87 epochs. 
The results of all the downstream tasks are listed in Table \ref{tab:all}.

\paragraph{NLI} 
Among the large models, {\GottBERT{large}{x}} achieved an accuracy of 82.46\%, while {\GottBERTf{large}{best}} and {\GottBERTf{large}{last}} slightly improved on this with accuracies of 83.31\% and 82.79\%, respectively. These results position the GottBERT models as strong contenders, though they were outperformed by GELECTRA\textsubscript{large}, which achieved the highest accuracy of 86.33\%. GBERT\textsubscript{large} also performed well with an accuracy of 84.21\%, followed closely by XLM-R\textsubscript{large} with 84.07\%.

For the base models, {\GottBERT{base}{best}} and {\GottBERT{base}{last}} achieved accuracies of 80.82\% and 81.04\%, respectively, demonstrating competitive performance. {\GottBERTf{base}{best}} and {\GottBERTf{base}{last}} had similar accuracies of 80.56\% and 80.74\%, respectively. Among the base models, GELECTRA\textsubscript{base} outperformed the others with an accuracy of 81.70\%. GBERT\textsubscript{base} scored slightly lower with 80.06. Other models like GermanBERT and XLM-R\textsubscript{base} achieved 78.16\% and 79.76\%, respectively, while dbmdzBERT and mBERT had the lowest accuracies at 68.12\% and 77.03\%.

Overall, the results indicate that while the {\GottBERTt} models exhibit strong performance in the NLI task, GELECTRA models generally achieved the highest accuracies in both the base and large categories.

\begin{table*}[h!tbp]
\begin{center}
\begin{tabular}{lccccccc}%
    \hline
    \multirow{2}{*}{\bfseries Model} & \multirow{2}{*}{\bfseries XNLI} & \multirow{2}{*}{\bfseries GermEval 2014} & \multirow{2}{*}{\bfseries CoNLL 03} & \multicolumn{2}{c}{\bfseries GermEval 2018} & \multirow{2}{*}{\bfseries 10kGNAD} \\
    &  &  & & \bfseries coarse & \bfseries fine &  
    \\\hline
    \csvreader[late after line = \\]{all_base.csv}{}
     {\csvcoli\ & \csvcolii & \csvcolv & \csvcolviii & \csvcolxi & \csvcolxiv & \csvcolxvii}
     \hline
    \csvreader[late after line = \\]{all_large.csv}{}
     {\csvcoli\ & \csvcolii & \csvcolv & \csvcolviii & \csvcolxi & \csvcolxiv & \csvcolxvii}
     \hline
     
\end{tabular}
\caption{\label{tab:all}All the results of the experiments are shown in percent. They are all based on the test set and the best score out of 24 runs (selection based on validation set). While NLI is measured by accuracy, all the other metrics are $F_1$ measures. Best score in bold and second underlined, for large and base models respectively.}
\end{center}
\end{table*}

\paragraph{Named Entity Recognition}

For the NER tasks, the base versions of the {\GottBERTt} models showed competitive performance with F1 scores around 87.50\% on the GermEval 2014 dataset and around 86.10\% on the CoNLL dataset. The large versions of these models improved upon these scores, with \GottBERTf{last}{x} achieving an F1 score of 88.27\% on GermEval 2014 and 86.78\% on CoNLL. However, among the large models, XLM-R achieved the highest F1 score of 88.83 on the GermEval 2014 dataset, whereas GBERT\textsubscript{large} performed the best on the CoNLL dataset with an F1 score of 87.19\%. Overall, the large {\GottBERTt} models demonstrated robust performance across both datasets, validating their effectiveness for the tasks at hand. Among the base architecture the {\GottBERTt} models took the lead.

\paragraph{Text Classification}

For GermEval 2018, the large {\GottBERTt} models showed again competitive performance. The {\GottBERT{large}{x}} and {\GottBERTf{large}{best}} models achieved overall F1 scores of around 79.3 for coarse-grained predictions, with minimal differences in fine-grained scores around 54.7. {\GottBERTf{large}{last}} had slightly lower performance in coarse predictions but was consistent in fine-grained predictions. In comparison, GELECTRA\textsubscript{large} outperformed all large models in coarse-grained predictions with an F1 score of 81.28, and also showed strong fine-grained performance with an F1 score of 56.17. GBERT\textsubscript{large} followed closely with 80.84 in coarse-grained predictions and led in fine-grained predictions with 57.37. XLM-R\textsubscript{large} scored slightly lower than the {\GottBERTt} models, with 79.05 and 55.06 in coarse and fine-grained predictions, respectively.

Among the base models, {\GottBERT{base}{best}} scored 78.17 for coarse-grained predictions and 53.30 for fine-grained predictions. {\GottBERT{base}{last}} showed similar performance in coarse-grained prediction with an F1 score of 78.18 and a fine-grained F1 score of 53.92. {\GottBERTf{base}{best}} achieved the highest coarse-grained F1 score of 78.65 but had a lower fine-grained score of 52.82. GELECTRA\textsubscript{base} scored 77.26 and 50.07 for coarse and fine-grained predictions and therefore scored close to XLM-R\textsubscript{base}. GBERT\textsubscript{base} and dbmdzBERT demonstrated moderate performance, while GermanBERT and mBERT had the lowest scores.

The evaluation of GottBERT models on the 10kGNAD dataset demonstrated their strong performance in German news classification tasks. For the large models, {\GottBERT{large}{x}} achieved an accuracy of 90.24, while {\GottBERTf{large}{best}} and {\GottBERTf{large}{last}} scored 90.31 and 90.17, respectively. Among the competing models, GELECTRA\textsubscript{large} outperformed all with an accuracy of 90.97, followed by GBERT\textsubscript{large} at 90.74, and XLM-R\textsubscript{large} matching {\GottBERTf{large}{last}} at 90.17.

For the base models, {\GottBERT{base}{last}} excelled with an accuracy of 90.27, while {\GottBERT{base}{best}} and {\GottBERTf{base}{best}} achieved 89.64 and 89.79, respectively. {\GottBERTf{base}{last}} performed well with an accuracy of 89.92. Notably, dbmdzBERT scored the highest among the base models with 90.34, followed closely by GBERT\textsubscript{base} at 90.30. GermanBERT, XLM-R\textsubscript{base}, and mBERT also showed competitive accuracies ranging from 88.90 to 90.18.

Overall, GottBERT models demonstrate strong and consistent performance across the classification tasks, highlighting their robustness and effectiveness.

\section{Discussion}
In this study, we successfully trained and evaluated {\GottBERTt} models on two versions of the OSCAR corpus.
Noteworthily, \citet{scheible2020gottbert} published {\GottBERT{base}{last}} as preliminary work. Since its release, the model has been utilized for various purposes in several related works showing its relevance. It has served as a baseline model in research studies \cite{scherrmann2023german, BRESSEM2024121598, 10.1093/jamiaopen/ooac087}. Beyond that, in the field of neural machine translation (NMT), researchers have used contextualized embeddings from pre-trained models including {\GottBERTt}\cite{xu2021bert}. Additionally, the model has been applied to named entity recognition (NER) tasks in the medical field, using both translated \cite{frei2022gernermed} and synthetic \cite{FREI2023104478} NER data annotated with medical entities through a fine-tuned version of GottBERT. Furthermore, a specialized version of the model known as BioGottBERT has been developed specifically for the medical domain \cite{10.1093/jamiaopen/ooac087}.

TPU training generally does not permit dynamic memory allocation, as TPUs are designed for efficient, high-throughput computation with fixed memory allocation. As a result, the corpus was processed as a single stream rather without considering document boundaries, unlike RoBERTa training on GPUs. Additionally, due to limitations in the fairseq implementation we used, we conducted the computations in 32-bit mode since 16-bit was neither properly implemented nor tested, leading to increased memory usage and hence more computation time required. Also, we used more conservative learning rates than the ones recommended by the fairseq documentation for pre-training on GPU.

Dataset annotation is expensive, as it is usually performed by multiple annotators. As the one-class SVD is trained unsupervised and annotations were only used to estimate its performance and to find a good $nu$, the dataset was only annotated by one person of our team. The MCC of 0.5663 suggests the model has a moderate ability to make accurate binary predictions overall, balancing true and false positives and negatives. The F1-score of 0.8578 indicates the model is performing well in terms of precision and recall for the positive class. These metrics together imply that while the model is quite good at correctly identifying positive cases and maintaining a balance between precision and recall, there is still room for improvement in making more accurate overall predictions as reflected in the moderate MCC score. Possibly a better approach would have been possible with a more complex model, the computational cost as well as the annotation efforts would have been much more expensive.
However, the use of language properties on a syntactical level denoted an efficient and creative approach that could be carried out with fair effort, including in terms of computational costs.

Our results do not provide a clear conclusion regarding the impact of data cleaning on the resulting model. A ranking of all base models by their position, taking into account only the number of first and second place models, shows that {\GottBERTf{base}{best}} is on top. However, when considering the {\GottBERTt} models in isolation as a subgroup, the {\GottBERT{base}{last}} model stands out as the top performer among the base models. Conversely, when evaluating the large models all {\GottBERTt} models were outperformed by the competitors. In this global comparison, the unfiltered model {\GottBERT{large}{best}} emerges as the best performer of all the {\GottBERTt} models winning one second place in CoNLL03, while {\GottBERTf{large}{best}} emerges as the superior performer in the isolated comparison. We anticipated a more definitive outcome, particularly since the filtered models were pre-trained for an additional epoch due to the smaller corpus size. The importance of hyperparameters in model performance is well-documented, even considering random seeds as shown by \citet{dodge_fine-tuning_2020}. This suggests that our chosen hyperparameters could be extended even more to find better ones. Nevertheless, clear differences should have been already pointed out within our experimental setup. However, we did not see any great benefit, especially considering the high cost of cleaning a data set in this way.

Potentially, the data cleaning process might have inadvertently removed important variance from the corpus. According to the \citet{martin_camembert_2020}, a corpus with greater variance generally leads to better performance compared to a homogeneous one. Therefore, we suggest creating a corpus with more variance. In our case, incorporating additional corpora such as OPUS, Wikipedia, and OpenLegalData could have been beneficial. Moreover, whole word masking (WWM) leads to better models \cite{martin_camembert_2020, chan-etal-2020-germans}. Finally, for RoBERTa models, the size of the vocabulary also impacts performance, as investigated by \citet{10.1145/3578707}. According to their findings, our vocabulary size wasn't a bad choice.

Finally, potential risks include bias and fairness issues, leading to unfair outcomes. Data privacy concerns exist, with the model potentially revealing sensitive information. This affects especially the corpus used and the filtered version of it, as the filtering did not operate on a semantical but on a syntactical level. Misuse could result in harmful content, like misinformation or spam. Over-reliance without human oversight might cause critical errors, especially in healthcare or finance. The environmental impact of training such models is considerable due to high energy consumption. It is also vulnerable to adversarial attacks.

\section{Conclusion}
In this work we present the German single language RoBERTa based model {\GottBERTt} in two versions computed on a corpus with 145GB and a filtered version with 121GB plain text with both base and large RoBERTa architecture. {\GottBERTt} is the first German single language RoBERTa based model.
In our experiments, we were able to show that the base models lead 4 of 6 tasks. However, this did not apply to the large models. The comparison of the pre-training with filtered and raw corpus did not show a clear result as anticipated. We therefore suggest considering other measures, such as increasing variance by using many corpora and using WWM. We release all {\GottBERTt} models in Huggingface and fairseq format to the community under the MIT license.

\section*{Acknowledgments}
This work was supported by the German Ministry for Education and Research (BMBF FKZ 01ZZ1801B, 01ZZ1804A, 01KX2121, and 01ZZ2304A) and supported with Cloud TPUs from Google's TPU Research Cloud (TRC). We would like to thank Ian Graham for constructive criticism of the manuscript and Louis Martin for the helping email contact. A special thanks goes to Myle Ott, who implemented the TPU feature in fairseq and intensively supported us to get our computation run. Finally, we would like to recognizably thank the people behind the scenes who essentially made this work possible: Frank Werner, Georg Koch and Friedlinde Bühler of the IMBI in Freiburg, Andreas Enterrottacher of the AIIM team in Munich, Philipp Munz and Christian Wiedemann of Wabion GmbH, Carsten Peters of Digital Schooling UG and last but not least Nora Limbourg the Google Cloud Customer Engineer assigned to us.

\section*{Limitations}
Several limitations need to be acknowledged in this study. First, the cleaning algorithm used was simple yet fast, but it could still be enhanced. Improved cleaning methods might lead to a clearer picture.

Further, the results obtained are specific to the first version of the OSCAR corpus, and this perspective may not generalize to other corpora or different languages. The performance of the models might vary significantly when applied to different datasets. Our evaluation tasks have already indicated this.
The models may struggle with cultural nuances and dialects within German. While {\GottBERTt} performs well on the used benchmarks, it may not adapt to evolving language use without further training.

Lastly, due to limited resources, we did not experiment with various learning rates (LR) for the pre-training of the large models. We chose a conservative peak LR of 0.00015, assuming that other learning rates could potentially lead to significantly better performance. However, given that large models require ca. 4.75 times more computation time, it was not feasible for us to explore different learning rates extensively.


\bibliography{custom}

@article{wolf_huggingfaces_2020,
  author    = {Thomas Wolf and
               Lysandre Debut and
               Victor Sanh and
               Julien Chaumond and
               Clement Delangue and
               Anthony Moi and
               Pierric Cistac and
               Tim Rault and
               R{\'{e}}mi Louf and
               Morgan Funtowicz and
               Jamie Brew},
  title     = {HuggingFace's Transformers: State-of-the-art Natural Language Processing},
  journal   = {CoRR},
  volume    = {abs/1910.03771},
  year      = {2019},
  url       = {http://arxiv.org/abs/1910.03771},
  archivePrefix = {arXiv},
  eprint    = {1910.03771},
  bibsource = {dblp computer science bibliography, https://dblp.org}
}

@article{benikova_germeval_2014,
	title = {{GermEval} 2014 {Named} {Entity} {Recognition} {Shared} {Task}: {Companion} {Paper}},
	language = {en},
	journal={Proceedings of the KONVENS GermEval Shared Task on Named Entity Recognition},
    pages={104--112},
	author = {Benikova, Darina and Biemann, Chris and Kisselew, Max and Padó, Sebastian},
	year = {2014},
}

@article{delobelle_robbert_2020,
	title = {{RobBERT}: a {Dutch} {RoBERTa}-based {Language} {Model}},
	shorttitle = {{RobBERT}},
	url = {http://arxiv.org/abs/2001.06286},
	urldate = {2020-02-17},
	journal = {arXiv:2001.06286 [cs]},
	author = {Delobelle, Pieter and Winters, Thomas and Berendt, Bettina},
	month = jan,
	year = {2020},
	note = {arXiv: 2001.06286}
}

@article{liu_roberta_2019,
	title = {{RoBERTa}: {A} {Robustly} {Optimized} {BERT} {Pretraining} {Approach}},
	shorttitle = {{RoBERTa}},
	url = {http://arxiv.org/abs/1907.11692},
	urldate = {2020-02-17},
	journal = {arXiv:1907.11692 [cs]},
	author = {Liu, Yinhan and Ott, Myle and Goyal, Naman and Du, Jingfei and Joshi, Mandar and Chen, Danqi and Levy, Omer and Lewis, Mike and Zettlemoyer, Luke and Stoyanov, Veselin},
	month = jul,
	year = {2019},
	note = {arXiv: 1907.11692}
}

@article{de_vries_bertje_2019,
	title = {{BERTje}: {A} {Dutch} {BERT} {Model}},
	shorttitle = {{BERTje}},
	url = {http://arxiv.org/abs/1912.09582},
	urldate = {2020-03-02},
	journal = {arXiv:1912.09582 [cs]},
	author = {de Vries, Wietse and van Cranenburgh, Andreas and Bisazza, Arianna and Caselli, Tommaso and van Noord, Gertjan and Nissim, Malvina},
	month = dec,
	year = {2019},
	note = {arXiv: 1912.09582}
}

@inproceedings{kudo_sentencepiece_2018,
	address = {Brussels, Belgium},
	title = {{SentencePiece}: {A} simple and language independent subword tokenizer and detokenizer for {Neural} {Text} {Processing}},
	shorttitle = {{SentencePiece}},
	url = {https://www.aclweb.org/anthology/D18-2012},
	doi = {10.18653/v1/D18-2012},
	urldate = {2020-04-03},
	booktitle = {Proceedings of the 2018 {Conference} on {Empirical} {Methods} in {Natural} {Language} {Processing}: {System} {Demonstrations}},
	publisher = {Association for Computational Linguistics},
	author = {Kudo, Taku and Richardson, John},
	month = nov,
	year = {2018},
	pages = {66--71}
}

@article{you_large_2020,
	title = {Large {Batch} {Optimization} for {Deep} {Learning}: {Training} {BERT} in 76 minutes},
	shorttitle = {Large {Batch} {Optimization} for {Deep} {Learning}},
	url = {http://arxiv.org/abs/1904.00962},
	urldate = {2020-04-02},
	journal = {arXiv:1904.00962 [cs, stat]},
	author = {You, Yang and Li, Jing and Reddi, Sashank and Hseu, Jonathan and Kumar, Sanjiv and Bhojanapalli, Srinadh and Song, Xiaodan and Demmel, James and Keutzer, Kurt and Hsieh, Cho-Jui},
	month = jan,
	year = {2020},
	note = {arXiv: 1904.00962
version: 5}
}

@article{radford_language_2019,
	title = {Language models are unsupervised multitask learners},
	volume = {1},
	number = {8},
	journal = {OpenAI Blog},
	author = {Radford, Alec and Wu, Jeffrey and Child, Rewon and Luan, David and Amodei, Dario and Sutskever, Ilya},
	year = {2019},
	pages = {9}
}

@inproceedings{cavnar_n-gram-based_1994,
	title = {N-{Gram}-{Based} {Text} {Categorization}},
	booktitle = {In {Proceedings} of {SDAIR}-94, 3rd {Annual} {Symposium} on {Document} {Analysis} and {Information} {Retrieval}},
	author = {Cavnar, William B. and Trenkle, John M.},
	year = {1994},
	pages = {161--175}
}

@inproceedings{tjong_kim_sang_introduction_2003,
	address = {USA},
	series = {{CONLL} ’03},
	title = {Introduction to the {CoNLL}-2003 {Shared} {Task}: {Language}-{Independent} {Named} {Entity} {Recognition}},
	url = {https://doi.org/10.3115/1119176.1119195},
	doi = {10.3115/1119176.1119195},
	booktitle = {Proceedings of the {Seventh} {Conference} on {Natural} {Language} {Learning} at {HLT}-{NAACL} 2003 - {Volume} 4},
	publisher = {Association for Computational Linguistics},
	author = {Tjong Kim Sang, Erik F. and De Meulder, Fien},
	year = {2003},
	note = {event-place: Edmonton, Canada},
	pages = {142--147}
}

@inproceedings{risch_fine-grained_2018,
	title = {Fine-{Grained} {Classification} of {Offensive} {Language}},
	booktitle = {Proceedings of {GermEval} 2018 (co-located with {KONVENS})},
	author = {Risch, Julian and Krebs, Eva and Löser, Alexander and Riese, Alexander and Krestel, Ralf},
	month = sep,
	year = {2018},
	pages = {38--44}
}

@article{conneau_unsupervised_2019,
  author    = {Alexis Conneau and
               Kartikay Khandelwal and
               Naman Goyal and
               Vishrav Chaudhary and
               Guillaume Wenzek and
               Francisco Guzm{\'{a}}n and
               Edouard Grave and
               Myle Ott and
               Luke Zettlemoyer and
               Veselin Stoyanov},
  title     = {Unsupervised Cross-lingual Representation Learning at Scale},
  journal   = {CoRR},
  volume    = {abs/1911.02116},
  year      = {2019},
  url       = {http://arxiv.org/abs/1911.02116},
  archivePrefix = {arXiv},
  eprint    = {1911.02116},
  bibsource = {dblp computer science bibliography, https://dblp.org}
}

@inproceedings{NEURIPS2019_c04c19c2,
 author = {Conneau, Alexis and Lample, Guillaume},
 booktitle = {Advances in Neural Information Processing Systems},
 editor = {H. Wallach and H. Larochelle and A. Beygelzimer and F. d\textquotesingle Alch\'{e}-Buc and E. Fox and R. Garnett},
 pages = {},
 publisher = {Curran Associates, Inc.},
 title = {Cross-lingual Language Model Pretraining},
 url = {https://proceedings.neurips.cc/paper_files/paper/2019/file/c04c19c2c2474dbf5f7ac4372c5b9af1-Paper.pdf},
 volume = {32},
 year = {2019}
}

@article{virtanen_multilingual_2019,
	title = {Multilingual is not enough: {BERT} for {Finnish}},
	shorttitle = {Multilingual is not enough},
	url = {http://arxiv.org/abs/1912.07076},
	urldate = {2020-05-25},
	journal = {arXiv:1912.07076 [cs]},
	author = {Virtanen, Antti and Kanerva, Jenna and Ilo, Rami and Luoma, Jouni and Luotolahti, Juhani and Salakoski, Tapio and Ginter, Filip and Pyysalo, Sampo},
	month = dec,
	year = {2019},
	note = {arXiv: 1912.07076}
}

@article{ott_fairseq_2019,
	title = {fairseq: {A} {Fast}, {Extensible} {Toolkit} for {Sequence} {Modeling}},
	shorttitle = {fairseq},
	url = {http://arxiv.org/abs/1904.01038},
	urldate = {2020-05-23},
	journal = {arXiv:1904.01038 [cs]},
	author = {Ott, Myle and Edunov, Sergey and Baevski, Alexei and Fan, Angela and Gross, Sam and Ng, Nathan and Grangier, David and Auli, Michael},
	month = apr,
	year = {2019},
	note = {arXiv: 1904.01038}
}

@misc{solling_sma_2009,
	title = {Små bokstäver ökade avståndet till tyskarna},
	url = {https://spraktidningen.se/artiklar/2009/06/sma-bokstaver-okade-avstandet-till-tyskarna},
	language = {sv},
	urldate = {2020-05-23},
	journal = {Språktidningen},
	author = {Solling, Daniel},
	month = jun,
	year = {2009},
	note = {Library Catalog: spraktidningen.se}
}

@book{crystal_cambridge_2003,
	title = {The {Cambridge} {Encyclopedia} of the {English} {Language}},
	isbn = {978-0-521-53033-0},
	language = {en},
	publisher = {Cambridge University Press},
	author = {Crystal, David and Crystal, Honorary Professor of Linguistics David},
	month = aug,
	year = {2003}
}

@article{ott_scaling_2018,
	title = {Scaling {Neural} {Machine} {Translation}},
	url = {http://arxiv.org/abs/1806.00187},
	urldate = {2020-05-21},
	journal = {arXiv:1806.00187 [cs]},
	author = {Ott, Myle and Edunov, Sergey and Grangier, David and Auli, Michael},
	month = sep,
	year = {2018},
	note = {arXiv: 1806.00187}
}

@article{le_flaubert_2020,
	title = {{FlauBERT}: {Unsupervised} {Language} {Model} {Pre}-training for {French}},
	shorttitle = {{FlauBERT}},
	url = {http://arxiv.org/abs/1912.05372},
	urldate = {2020-06-16},
	journal = {arXiv:1912.05372 [cs]},
	author = {Le, Hang and Vial, Loïc and Frej, Jibril and Segonne, Vincent and Coavoux, Maximin and Lecouteux, Benjamin and Allauzen, Alexandre and Crabbé, Benoît and Besacier, Laurent and Schwab, Didier},
	month = mar,
	year = {2020},
	note = {arXiv: 1912.05372}
}

@inproceedings{koehn_moses_2007,
	address = {Prague, Czech Republic},
	title = {Moses: {Open} {Source} {Toolkit} for {Statistical} {Machine} {Translation}},
	url = {https://www.aclweb.org/anthology/P07-2045},
	booktitle = {Proceedings of the 45th {Annual} {Meeting} of the {Association} for {Computational} {Linguistics} {Companion} {Volume} {Proceedings} of the {Demo} and {Poster} {Sessions}},
	publisher = {Association for Computational Linguistics},
	author = {Koehn, Philipp and Hoang, Hieu and Birch, Alexandra and Callison-Burch, Chris and Federico, Marcello and Bertoldi, Nicola and Cowan, Brooke and Shen, Wade and Moran, Christine and Zens, Richard and Dyer, Chris and Bojar, Ondřej and Constantin, Alexandra and Herbst, Evan},
	month = jun,
	year = {2007},
	pages = {177--180}
}

@inproceedings{schabus_one_2017,
	address = {Tokyo, Japan},
	title = {One {Million} {Posts}: {A} {Data} {Set} of {German} {Online} {Discussions}},
	doi = {10.1145/3077136.3080711},
	booktitle = {Proceedings of the 40th {International} {ACM} {SIGIR} {Conference} on {Research} and {Development} in {Information} {Retrieval} ({SIGIR})},
	author = {Schabus, Dietmar and Skowron, Marcin and Trapp, Martin},
	month = aug,
	year = {2017},
	pages = {1241--1244}
}

@inproceedings{martin_camembert_2020,
	address = {Online},
	title = {{CamemBERT}: a {Tasty} {French} {Language} {Model}},
	url = {https://www.aclweb.org/anthology/2020.acl-main.645},
	booktitle = {Proceedings of the 58th {Annual} {Meeting} of the {Association} for {Computational} {Linguistics}},
	publisher = {Association for Computational Linguistics},
	author = {Martin, Louis and Muller, Benjamin and Ortiz Suárez, Pedro Javier and Dupont, Yoann and Romary, Laurent and de la Clergerie, Éric and Seddah, Djamé and Sagot, Benoît},
	month = jul,
	year = {2020},
	pages = {7203--7219}
}

@inproceedings{devlin_bert_2019,
	address = {Minneapolis, Minnesota},
	title = {{BERT}: {Pre}-training of {Deep} {Bidirectional} {Transformers} for {Language} {Understanding}},
	url = {https://www.aclweb.org/anthology/N19-1423},
	doi = {10.18653/v1/N19-1423},
	booktitle = {Proceedings of the 2019 {Conference} of the {North} {American} {Chapter} of the {Association} for {Computational} {Linguistics}: {Human} {Language} {Technologies}, {Volume} 1 ({Long} and {Short} {Papers})},
	publisher = {Association for Computational Linguistics},
	author = {Devlin, Jacob and Chang, Ming-Wei and Lee, Kenton and Toutanova, Kristina},
	month = jun,
	year = {2019},
	pages = {4171--4186}
}

@incollection{vaswani_attention_2017,
	title = {Attention is {All} you {Need}},
	url = {http://papers.nips.cc/paper/7181-attention-is-all-you-need.pdf},
	booktitle = {Advances in {Neural} {Information} {Processing} {Systems} 30},
	publisher = {Curran Associates, Inc.},
	author = {Vaswani, Ashish and Shazeer, Noam and Parmar, Niki and Uszkoreit, Jakob and Jones, Llion and Gomez, Aidan N and Kaiser, Łukasz and Polosukhin, Illia},
	editor = {Guyon, I. and Luxburg, U. V. and Bengio, S. and Wallach, H. and Fergus, R. and Vishwanathan, S. and Garnett, R.},
	year = {2017},
	pages = {5998--6008}
}

@inproceedings{ortiz_suarez_monolingual_2020,
	address = {Online},
	title = {A {Monolingual} {Approach} to {Contextualized} {Word} {Embeddings} for {Mid}-{Resource} {Languages}},
	url = {https://www.aclweb.org/anthology/2020.acl-main.156},
	booktitle = {Proceedings of the 58th {Annual} {Meeting} of the {Association} for {Computational} {Linguistics}},
	publisher = {Association for Computational Linguistics},
	author = {Ortiz Suárez, Pedro Javier and Romary, Laurent and Sagot, Benoît},
	month = jul,
	year = {2020},
	pages = {1703--1714}
}

@article{dodge_fine-tuning_2020,
	title = {Fine-{Tuning} {Pretrained} {Language} {Models}: {Weight} {Initializations}, {Data} {Orders}, and {Early} {Stopping}},
	shorttitle = {Fine-{Tuning} {Pretrained} {Language} {Models}},
	url = {http://arxiv.org/abs/2002.06305},
	urldate = {2020-09-16},
	journal = {arXiv:2002.06305 [cs]},
	author = {Dodge, Jesse and Ilharco, Gabriel and Schwartz, Roy and Farhadi, Ali and Hajishirzi, Hannaneh and Smith, Noah},
	month = feb,
	year = {2020},
	note = {arXiv: 2002.06305}
}

@inproceedings{ng_facebook_2019,
	address = {Florence, Italy},
	title = {Facebook {FAIR}'s {WMT19} {News} {Translation} {Task} {Submission}},
	url = {https://www.aclweb.org/anthology/W19-5333},
	doi = {10.18653/v1/W19-5333},
	urldate = {2020-09-23},
	booktitle = {Proceedings of the {Fourth} {Conference} on {Machine} {Translation} ({Volume} 2: {Shared} {Task} {Papers}, {Day} 1)},
	publisher = {Association for Computational Linguistics},
	author = {Ng, Nathan and Yee, Kyra and Baevski, Alexei and Ott, Myle and Auli, Michael and Edunov, Sergey},
	month = aug,
	year = {2019},
	pages = {314--319}
}

@article{sennrich_neural_2016,
	title = {Neural {Machine} {Translation} of {Rare} {Words} with {Subword} {Units}},
	url = {http://arxiv.org/abs/1508.07909},
	urldate = {2020-09-24},
	journal = {arXiv:1508.07909 [cs]},
	author = {Sennrich, Rico and Haddow, Barry and Birch, Alexandra},
	month = jun,
	year = {2016},
	note = {arXiv: 1508.07909}
}

@inproceedings{schuster_japanese_2012,
	title = {Japanese and {Korean} voice search},
	doi = {10.1109/ICASSP.2012.6289079},
	booktitle = {2012 {IEEE} {International} {Conference} on {Acoustics}, {Speech} and {Signal} {Processing} ({ICASSP})},
	author = {Schuster, Mike and Nakajima, Kaisuke},
	month = mar,
	year = {2012},
	note = {ISSN: 2379-190X},
	pages = {5149--5152}
}

@misc{devlin_multilingual_2018,
	title = {Multilingual {BERT} {Readme} {Document}},
	url = {https://github.com/google-research/bert/blob/cc7051dc592802f501e8a6f71f8fb3cf9de95dc9/multilingual.md},
	language = {en},
	urldate = {2020-11-18},
	journal = {GitHub},
	author = {Devlin, Jacob},
	month = nov,
	year = {2018}
}

@inproceedings{scholkopf1999single,
  title={Single-class support vector machines},
  author={Sch{\"o}lkopf, B and Williamson, R and Smola, AJ and Shawe-Taylor, J},
  booktitle={Dagstuhl-Seminar 99121: Unsupervised Learning},
  pages={19--20},
  year={1999},
  organization={Schloss Dagstuhl, Leibniz-Zentrum f{\"u}r Informatik}
}

@misc{clark2020electra,
      title={ELECTRA: Pre-training Text Encoders as Discriminators Rather Than Generators}, 
      author={Kevin Clark and Minh-Thang Luong and Quoc V. Le and Christopher D. Manning},
      year={2020},
      eprint={2003.10555},
      archivePrefix={arXiv},
      primaryClass={cs.CL}
}

@misc{jouppi_tpu_2023,
	title = {{TPU} v4: An Optically Reconfigurable Supercomputer for Machine Learning with Hardware Support for Embeddings},
	url = {http://arxiv.org/abs/2304.01433},
	doi = {10.48550/arXiv.2304.01433},
	shorttitle = {{TPU} v4},
    year = {2023},
	number = {{arXiv}:2304.01433},
	publisher = {{arXiv}},
	author = {Jouppi, Norman P. and Kurian, George and Li, Sheng and Ma, Peter and Nagarajan, Rahul and Nai, Lifeng and Patil, Nishant and Subramanian, Suvinay and Swing, Andy and Towles, Brian and Young, Cliff and Zhou, Xiang and Zhou, Zongwei and Patterson, David},
	urldate = {2023-07-20},
	date = {2023-04-20},
	eprinttype = {arxiv},
	eprint = {2304.01433 [cs]},
}

@inproceedings{chan-etal-2020-germans,
    title = "{G}erman{'}s Next Language Model",
    author = {Chan, Branden  and
      Schweter, Stefan  and
      M{\"o}ller, Timo},
    editor = "Scott, Donia  and
      Bel, Nuria  and
      Zong, Chengqing",
    booktitle = "Proceedings of the 28th International Conference on Computational Linguistics",
    month = dec,
    year = "2020",
    address = "Barcelona, Spain (Online)",
    publisher = "International Committee on Computational Linguistics",
    url = "https://aclanthology.org/2020.coling-main.598",
    doi = "10.18653/v1/2020.coling-main.598",
    pages = "6788--6796",
}

@article{10.1145/3578707,
author = {Toraman, Cagri and Yilmaz, Eyup Halit and \c{S}ahinu\c{c} Furkan and Ozcelik, Oguzhan},
title = {Impact of Tokenization on Language Models: An Analysis for Turkish},
year = {2023},
issue_date = {April 2023},
publisher = {Association for Computing Machinery},
address = {New York, NY, USA},
volume = {22},
number = {4},
issn = {2375-4699},
url = {https://doi.org/10.1145/3578707},
doi = {10.1145/3578707},
journal = {ACM Trans. Asian Low-Resour. Lang. Inf. Process.},
month = {mar},
articleno = {116},
numpages = {21}
}

@article{chicco_advantages_2020,
	title = {The advantages of the {Matthews} correlation coefficient ({MCC}) over {F1} score and accuracy in binary classification evaluation},
	volume = {21},
	issn = {1471-2164},
	url = {https://doi.org/10.1186/s12864-019-6413-7},
	doi = {10.1186/s12864-019-6413-7},
	number = {1},
	urldate = {2023-04-28},
	journal = {BMC Genomics},
	author = {Chicco, Davide and Jurman, Giuseppe},
	month = jan,
	year = {2020},
	pages = {6},
}

@article{chicco_matthews_2021,
	title = {The {Matthews} {Correlation} {Coefficient} ({MCC}) is {More} {Informative} {Than} {Cohen}’s {Kappa} and {Brier} {Score} in {Binary} {Classification} {Assessment}},
	volume = {9},
	issn = {2169-3536},
	doi = {10.1109/ACCESS.2021.3084050},
	journal = {IEEE Access},
	author = {Chicco, Davide and Warrens, Matthijs J. and Jurman, Giuseppe},
	year = {2021},
	note = {Conference Name: IEEE Access},
	pages = {78368--78381},
}

@inproceedings{williams-etal-2018-broad,
    title = "A Broad-Coverage Challenge Corpus for Sentence Understanding through Inference",
    author = "Williams, Adina  and
      Nangia, Nikita  and
      Bowman, Samuel",
    editor = "Walker, Marilyn  and
      Ji, Heng  and
      Stent, Amanda",
    booktitle = "Proceedings of the 2018 Conference of the North {A}merican Chapter of the Association for Computational Linguistics: Human Language Technologies, Volume 1 (Long Papers)",
    month = jun,
    year = "2018",
    address = "New Orleans, Louisiana",
    publisher = "Association for Computational Linguistics",
    url = "https://aclanthology.org/N18-1101",
    doi = "10.18653/v1/N18-1101",
    pages = "1112--1122",
}

@inproceedings{conneau-etal-2018-xnli,
    title = "{XNLI}: Evaluating Cross-lingual Sentence Representations",
    author = "Conneau, Alexis  and
      Rinott, Ruty  and
      Lample, Guillaume  and
      Williams, Adina  and
      Bowman, Samuel  and
      Schwenk, Holger  and
      Stoyanov, Veselin",
    editor = "Riloff, Ellen  and
      Chiang, David  and
      Hockenmaier, Julia  and
      Tsujii, Jun{'}ichi",
    booktitle = "Proceedings of the 2018 Conference on Empirical Methods in Natural Language Processing",
    month = oct # "-" # nov,
    year = "2018",
    address = "Brussels, Belgium",
    publisher = "Association for Computational Linguistics",
    url = "https://aclanthology.org/D18-1269",
    doi = "10.18653/v1/D18-1269",
    pages = "2475--2485",
}

@misc{touvron2023llama1,
      title={LLaMA: Open and Efficient Foundation Language Models}, 
      author={Hugo Touvron and Thibaut Lavril and Gautier Izacard and Xavier Martinet and Marie-Anne Lachaux and Timothée Lacroix and Baptiste Rozière and Naman Goyal and Eric Hambro and Faisal Azhar and Aurelien Rodriguez and Armand Joulin and Edouard Grave and Guillaume Lample},
      year={2023},
      eprint={2302.13971},
      archivePrefix={arXiv},
      primaryClass={id='cs.CL' full_name='Computation and Language' is_active=True alt_name='cmp-lg' in_archive='cs' is_general=False description='Covers natural language processing. Roughly includes material in ACM Subject Class I.2.7. Note that work on artificial languages (programming languages, logics, formal systems) that does not explicitly address natural-language issues broadly construed (natural-language processing, computational linguistics, speech, text retrieval, etc.) is not appropriate for this area.'}
}

@misc{touvron2023llama2,
      title={Llama 2: Open Foundation and Fine-Tuned Chat Models}, 
      author={Hugo Touvron and Louis Martin and Kevin Stone and Peter Albert and Amjad Almahairi and Yasmine Babaei and Nikolay Bashlykov and Soumya Batra and Prajjwal Bhargava and Shruti Bhosale and Dan Bikel and Lukas Blecher and Cristian Canton Ferrer and Moya Chen and Guillem Cucurull and David Esiobu and Jude Fernandes and Jeremy Fu and Wenyin Fu and Brian Fuller and Cynthia Gao and Vedanuj Goswami and Naman Goyal and Anthony Hartshorn and Saghar Hosseini and Rui Hou and Hakan Inan and Marcin Kardas and Viktor Kerkez and Madian Khabsa and Isabel Kloumann and Artem Korenev and Punit Singh Koura and Marie-Anne Lachaux and Thibaut Lavril and Jenya Lee and Diana Liskovich and Yinghai Lu and Yuning Mao and Xavier Martinet and Todor Mihaylov and Pushkar Mishra and Igor Molybog and Yixin Nie and Andrew Poulton and Jeremy Reizenstein and Rashi Rungta and Kalyan Saladi and Alan Schelten and Ruan Silva and Eric Michael Smith and Ranjan Subramanian and Xiaoqing Ellen Tan and Binh Tang and Ross Taylor and Adina Williams and Jian Xiang Kuan and Puxin Xu and Zheng Yan and Iliyan Zarov and Yuchen Zhang and Angela Fan and Melanie Kambadur and Sharan Narang and Aurelien Rodriguez and Robert Stojnic and Sergey Edunov and Thomas Scialom},
      year={2023},
      eprint={2307.09288},
      archivePrefix={arXiv},
      primaryClass={cs.CL}
}

@article{Liu_2023,
   title={Summary of ChatGPT-Related research and perspective towards the future of large language models},
   volume={1},
   ISSN={2950-1628},
   url={http://dx.doi.org/10.1016/j.metrad.2023.100017},
   DOI={10.1016/j.metrad.2023.100017},
   number={2},
   journal={Meta-Radiology},
   publisher={Elsevier BV},
   author={Liu, Yiheng and Han, Tianle and Ma, Siyuan and Zhang, Jiayue and Yang, Yuanyuan and Tian, Jiaming and He, Hao and Li, Antong and He, Mengshen and Liu, Zhengliang and Wu, Zihao and Zhao, Lin and Zhu, Dajiang and Li, Xiang and Qiang, Ning and Shen, Dingang and Liu, Tianming and Ge, Bao},
   year={2023},
   month=sep, pages={100017} }

@misc{scheible2020gottbert,
      title={GottBERT: a pure German Language Model}, 
      author={Raphael Scheible and Fabian Thomczyk and Patric Tippmann and Victor Jaravine and Martin Boeker},
      year={2020},
      eprint={2012.02110},
      archivePrefix={arXiv},
      primaryClass={cs.CL}
}

@misc{scherrmann2023german,
      title={German FinBERT: A German Pre-trained Language Model}, 
      author={Moritz Scherrmann},
      year={2023},
      eprint={2311.08793},
      archivePrefix={arXiv},
      primaryClass={id='cs.CL' full_name='Computation and Language' is_active=True alt_name='cmp-lg' in_archive='cs' is_general=False description='Covers natural language processing. Roughly includes material in ACM Subject Class I.2.7. Note that work on artificial languages (programming languages, logics, formal systems) that does not explicitly address natural-language issues broadly construed (natural-language processing, computational linguistics, speech, text retrieval, etc.) is not appropriate for this area.'}
}

@article{BRESSEM2024121598,
title = {medBERT.de: A comprehensive German BERT model for the medical domain},
journal = {Expert Systems with Applications},
volume = {237},
pages = {121598},
year = {2024},
issn = {0957-4174},
doi = {https://doi.org/10.1016/j.eswa.2023.121598},
url = {https://www.sciencedirect.com/science/article/pii/S0957417423021000},
author = {Keno K. Bressem and Jens-Michalis Papaioannou and Paul Grundmann and Florian Borchert and Lisa C. Adams and Leonhard Liu and Felix Busch and Lina Xu and Jan P. Loyen and Stefan M. Niehues and Moritz Augustin and Lennart Grosser and Marcus R. Makowski and Hugo J.W.L. Aerts and Alexander Löser}
}

@misc{xu2021bert,
      title={BERT, mBERT, or BiBERT? A Study on Contextualized Embeddings for Neural Machine Translation}, 
      author={Haoran Xu and Benjamin Van Durme and Kenton Murray},
      year={2021},
      eprint={2109.04588},
      archivePrefix={arXiv},
      primaryClass={id='cs.CL' full_name='Computation and Language' is_active=True alt_name='cmp-lg' in_archive='cs' is_general=False description='Covers natural language processing. Roughly includes material in ACM Subject Class I.2.7. Note that work on artificial languages (programming languages, logics, formal systems) that does not explicitly address natural-language issues broadly construed (natural-language processing, computational linguistics, speech, text retrieval, etc.) is not appropriate for this area.'}
}

@article{10.1093/jamiaopen/ooac087,
    author = {Lentzen, Manuel and Madan, Sumit and Lage-Rupprecht, Vanessa and Kühnel, Lisa and Fluck, Juliane and Jacobs, Marc and Mittermaier, Mirja and Witzenrath, Martin and Brunecker, Peter and Hofmann-Apitius, Martin and Weber, Joachim and Fröhlich, Holger},
    title = "{Critical assessment of transformer-based AI models for German clinical notes}",
    journal = {JAMIA Open},
    volume = {5},
    number = {4},
    pages = {ooac087},
    year = {2022},
    month = {11},
    issn = {2574-2531},
    doi = {10.1093/jamiaopen/ooac087},
    url = {https://doi.org/10.1093/jamiaopen/ooac087},
    eprint = {https://academic.oup.com/jamiaopen/article-pdf/5/4/ooac087/47042093/ooac087.pdf},
}

@article{FREI2023104478,
title = {Annotated dataset creation through large language models for non-english medical NLP},
journal = {Journal of Biomedical Informatics},
volume = {145},
pages = {104478},
year = {2023},
issn = {1532-0464},
doi = {https://doi.org/10.1016/j.jbi.2023.104478},
url = {https://www.sciencedirect.com/science/article/pii/S1532046423001995},
author = {Johann Frei and Frank Kramer}
}

@misc{frei2022gernermed,
      title={GERNERMED++: Transfer Learning in German Medical NLP}, 
      author={Johann Frei and Ludwig Frei-Stuber and Frank Kramer},
      year={2022},
      eprint={2206.14504},
      archivePrefix={arXiv},
      primaryClass={id='cs.CL' full_name='Computation and Language' is_active=True alt_name='cmp-lg' in_archive='cs' is_general=False description='Covers natural language processing. Roughly includes material in ACM Subject Class I.2.7. Note that work on artificial languages (programming languages, logics, formal systems) that does not explicitly address natural-language issues broadly construed (natural-language processing, computational linguistics, speech, text retrieval, etc.) is not appropriate for this area.'}
}

@misc{jiang2023mistral,
      title={Mistral 7B}, 
      author={Albert Q. Jiang and Alexandre Sablayrolles and Arthur Mensch and Chris Bamford and Devendra Singh Chaplot and Diego de las Casas and Florian Bressand and Gianna Lengyel and Guillaume Lample and Lucile Saulnier and Lélio Renard Lavaud and Marie-Anne Lachaux and Pierre Stock and Teven Le Scao and Thibaut Lavril and Thomas Wang and Timothée Lacroix and William El Sayed},
      year={2023},
      eprint={2310.06825},
      archivePrefix={arXiv},
      primaryClass={id='cs.CL' full_name='Computation and Language' is_active=True alt_name='cmp-lg' in_archive='cs' is_general=False description='Covers natural language processing. Roughly includes material in ACM Subject Class I.2.7. Note that work on artificial languages (programming languages, logics, formal systems) that does not explicitly address natural-language issues broadly construed (natural-language processing, computational linguistics, speech, text retrieval, etc.) is not appropriate for this area.'}
}

\appendix

\section{Filtering OSCAR}
\label{sec:OSCAR}
\begin{figure}[!h]
    \centering
    \begin{subfigure}[b]{0.47\textwidth}
        \includegraphics[width=\columnwidth]{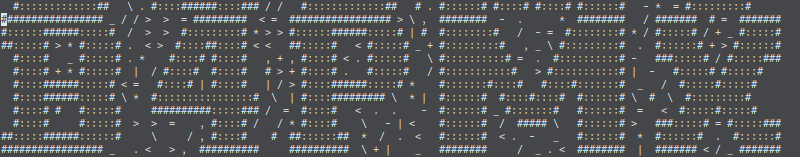}
        \caption{B4PMX written in ASCII.}
        \label{fig:ascii_font}
    \end{subfigure}

    \begin{subfigure}[b]{0.47\textwidth}
        \centering
        \includegraphics[width=\columnwidth]{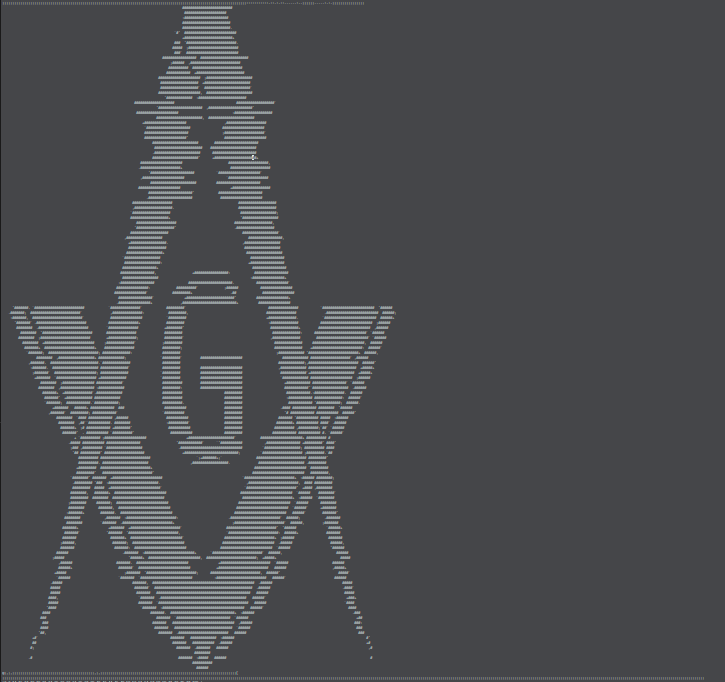}
        \caption{Freemasonry sign.}
        \label{fig:ascii_freemasonry}
    \end{subfigure}

    \begin{subfigure}[b]{0.47\textwidth}
        \centering
        \includegraphics[width=\columnwidth]{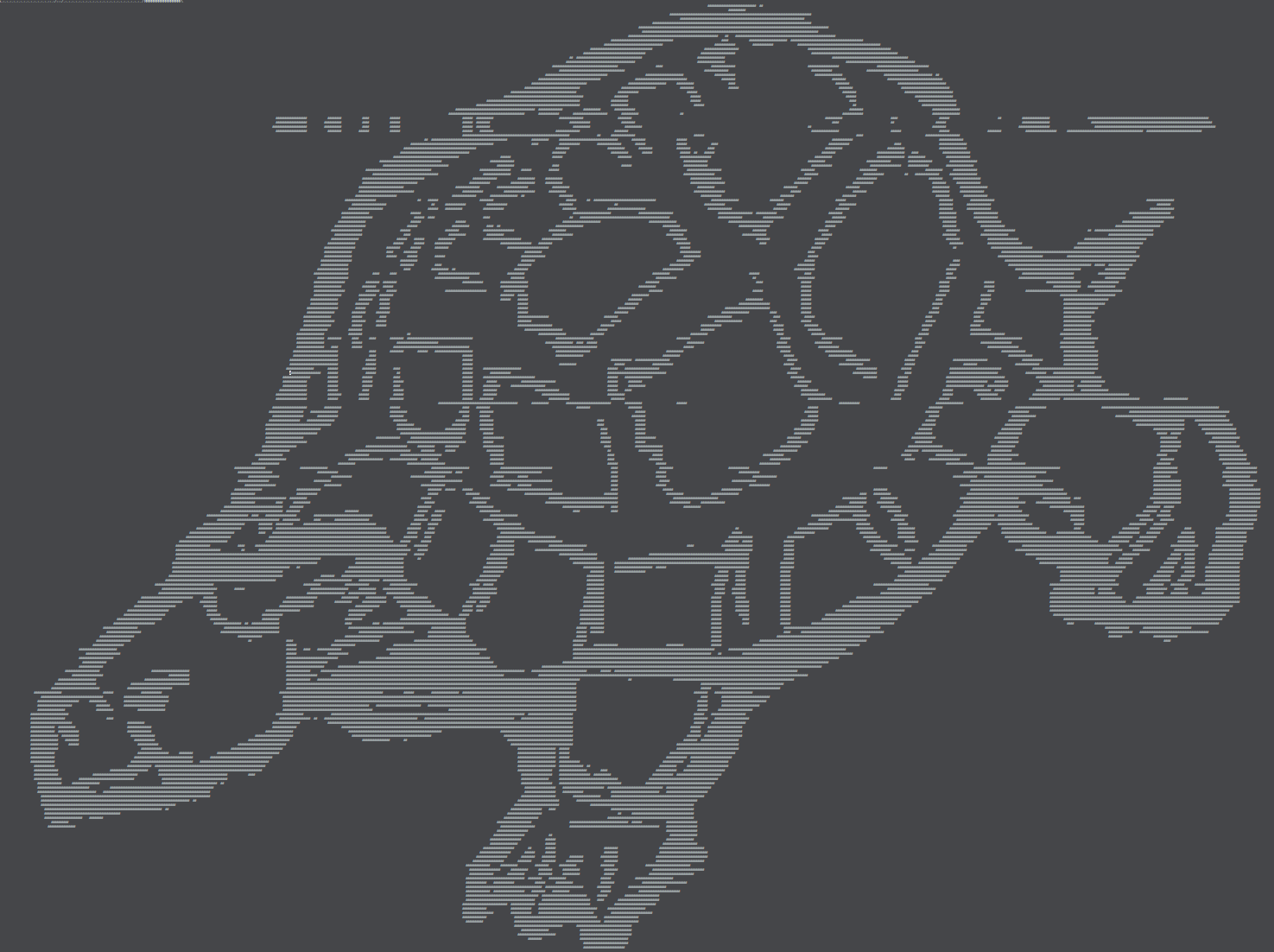}
        \caption{A turtle.}
        \label{fig:ascii_turtle}
    \end{subfigure}

    \caption{\label{fig:ascii_arts}ASCII arts found in OSCAR.}
\end{figure}

Inside OSCAR interesting artefacts were found. Besides encoding errors, mainly affecting Umlauts, there were occurrences of ASCII arts (see Figures \ref{fig:ascii_arts}) and source code.

\newpage

\section{Model Properties}
The number of parameters in BERT-like models varies significantly based on their architecture (see \ref{tab:params}). The base version of BERT has approximately 110 million parameters, while the large version has about 340 million. RoBERTa, an optimized version of BERT, has 125 million parameters in its base model and 355 million in the large model, benefiting from extended training and larger datasets. The multilingual XLM-RoBERTa comes in two main versions: the base model with around 270 million parameters and the large model with about 550 million, which helps handle multiple languages effectively. Electra, using a generator-discriminator framework, achieves high performance with fewer parameters, with the base model having about 110 million parameters and the large model around 335 million.
\begin{table}[h!tbp]
\begin{center}
\begin{tabular}{lcc}%
    \hline
    \bfseries Model & \bfseries Vocab Size & \bfseries \#Params
    \\\hline
    \csvreader[late after line = \\]{params.csv}{}
     {\csvcoli\ & \csvcolii & \csvcoliii}
     \hline
\end{tabular}
\caption{\label{tab:params}The size of the vocabulary and the size of the parameters are shown for the model types used in this study. This table does not show other design differences of the models. Values were extracted using Huggingface's transformers library.}
\end{center}
\end{table}

\section{Perplexity}
\label{sec:perplexity}

\begin{figure}[htb]
    \centering
    \begin{subfigure}[b]{0.5\textwidth}
         \centering
        \includegraphics[width=\columnwidth]{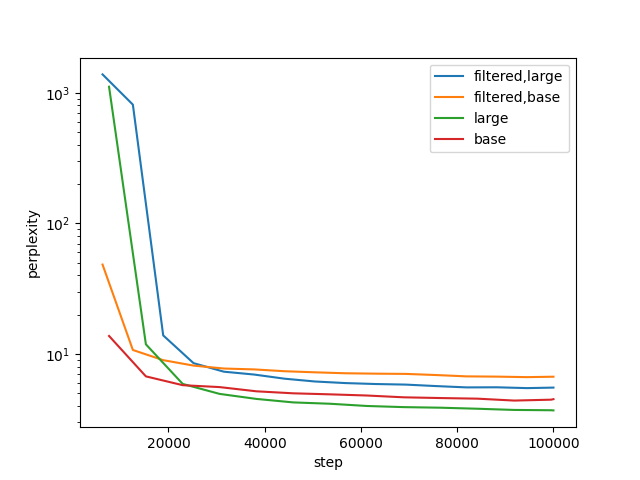}
    \end{subfigure}
    \begin{subfigure}[b]{0.5\textwidth}
         \centering
         \includegraphics[width=\columnwidth]{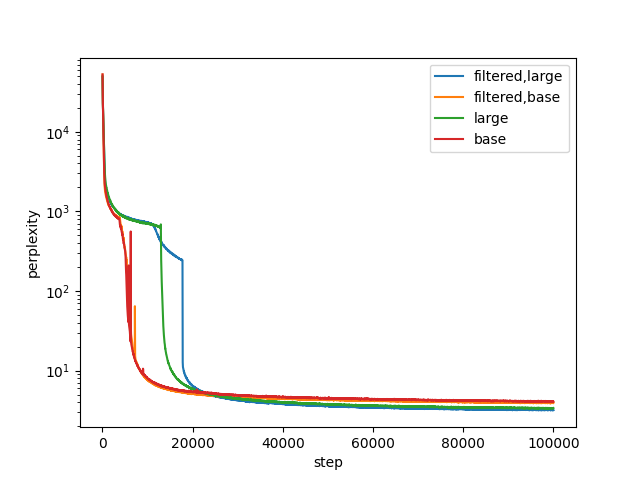}
    \end{subfigure}
    \caption{\label{fig:perplexity}Perplexity of the {\GottBERTt} models. Top based on a validation at the checkpoints. Bottom based on the validation of each optimization cycle during the training.}
\end{figure}

During the model pre-training the perplexity of the model is computed based on a test set for each optimization cycle and based on a validation set at each checkpoint (see Figure \ref{fig:perplexity}). Within the training all the curves show a plateau: the base models only a short one, while the large models have a relatively long one. Some models even have upward spikes, which could be interpreted locally as divergence when observing the training process if they are not known. Furthermore, we see a very flat convergence of the models after 40k steps at the latest. This convergence can also be seen within the validation set based perplexity which was computed at each epoch.

\newpage
\section{Parameters}
The parameter space for our grid search is listed in Table \ref{tab:hyperparams}. In addition, Table \ref{tab:best_params} shows the parameters of the best models (selection based on validation set) of the respective tasks. The time required for the evaluation is shown in Table 3. The tasks were computed on Nvidia Titan RTX and Nvidia A40 graphics devices and we relied on Huggingface's transformers library in version v4.34.1.

\begin{table}[htb]
    \centering
    \begin{tabular}{lc}
         \bfseries Parameter & \bfseries Values\\
         \hline
         Learning Rate & 5e-5, 2e-5, 1e-5, 7e-6, 5e-6, 1e-6 \\
         Batch Size & 16, 32, 48, 64 \\
         Epochs & 30 \\
         \hline
    \end{tabular}
    \caption{Hyperparameters used in the grid search of the downstream tasks.}
    \label{tab:hyperparams}
\end{table}

\begin{table*}[h!tbp]
\begin{center}
\begin{tabular}{lcccccccccc}%
    \hline
    \multirow{2}{*}{\bfseries Model} & 
    \multicolumn{2}{c}{\multirow{2}{*}{\bfseries GermEval 2014}} & 
    \multicolumn{2}{c}{\multirow{2}{*}{\bfseries CoNLL 03}} & 
    \multicolumn{4}{c}{\bfseries GermEval 2018} & 
    \multicolumn{2}{c}{\multirow{2}{*}{\bfseries 10kGNAD}} \\
    & & & & & \multicolumn{2}{c}{\bfseries coarse} & \multicolumn{2}{c}{\bfseries fine} & &  
    \\\hline
      & BF & LR & BF & LR & BF & LR & BF & LR & BF & LR 
    \\\cmidrule{2-11}
    \csvreader[late after line = \\]{hyperparams_base.csv}{}
     {\csvcoli & \csvcolii & \csvcoliii & \csvcoliv & \csvcolv & \csvcolvi & \csvcolvii & \csvcolviii & \csvcolix & \csvcolx & \csvcolxi}
     \hline
    \csvreader[late after line = \\]{hyperparams_large.csv}{}
     {\csvcoli & \csvcolii & \csvcoliii & \csvcoliv & \csvcolv & \csvcolvi & \csvcolvii & \csvcolviii & \csvcolix & \csvcolx & \csvcolxi}
     \hline
\end{tabular}
\caption{\label{tab:best_params}Hyperparameters of the best downstream task model of the respective tasks and pre-trained models. BF is the batch size and LR the learning rate.}
\end{center}
\end{table*}


\begin{table}[htb]
    \centering
    \begin{tabular}{llc}
         \hline
         \bfseries Task & & \bfseries Computation Time\\
         \hline
         \multicolumn{2}{l}{XNLI} & 672:59 \\
         \multicolumn{2}{l}{GermEval 2014} & 284:20 \\
         \multicolumn{2}{l}{CoNLL03} & 169:26 \\
         \multirow{2}{*}{GermEval 2018} & coarse & 113:36 \\
          & fine & 113:47 \\
         \multicolumn{2}{l}{10kGNAD} & 195:21 \\
         \hline
    \end{tabular}
    \caption{Computation time in hours and minutes for the downstream tasks summing up to 1549 hours and 29 minutes which are approximately 64.6 days.}
    \label{tab:gpu_time}
\end{table}

\end{document}